\documentclass[review]{elsarticle}

\usepackage{amsmath,amssymb,amsfonts,bm}
\usepackage{caption}
\usepackage{bm}
\usepackage{multirow}
\usepackage{subfigure}
\usepackage{booktabs}
\usepackage[ruled]{algorithm2e}

\usepackage[colorlinks, linkcolor=blue, anchorcolor=blue, citecolor=blue]{hyperref}
\usepackage{geometry}
\begin{document}

\begin{frontmatter}

\title{Physics-informed MTA-UNet: Prediction of Thermal Stress and Thermal Deformation of Satellites}

\author[mymainaddress,mysecondaryaddress]{Zeyu Cao}
\ead{Komorebi0216@163.com}

\author[mymainaddress]{Wen Yao\corref{mycorrespondingauthor}}
\cortext[mycorrespondingauthor]{Corresponding author}

\author[mymainaddress]{Wei Peng}

\author[mymainaddress]{Xiaoya Zhang}

\author[mysecondaryaddress]{Kairui Bao}

\address[mymainaddress]{Defense Innovation Institute, Chinese Academy of Military Science, No. 53, Fengtai East Street, Beijing 100071, China}

\address[mysecondaryaddress]{The No.92941st Troop of PLA, Huludao 125001, China}

\begin{abstract}
The rapid analysis of thermal stress and deformation plays a pivotal role in the thermal control measures and optimization of the structural design of satellites. For achieving real-time thermal stress and thermal deformation analysis of satellite motherboards, this paper proposes a novel Multi-Task Attention UNet (MTA-UNet) neural network which combines the advantages of both Multi-Task Learning (MTL) and U-Net with attention mechanism. Besides, a physics-informed strategy is used in the training process, where partial differential equations (PDEs) are integrated into the loss functions as residual terms. 
Finally, an uncertainty-based loss balancing approach is applied to weight different loss functions of multiple training tasks. Experimental results show that the proposed MTA-UNet effectively improves the prediction accuracy of multiple physics tasks compared with Single-Task Learning (STL) models. In addition, the physics-informed method brings less error in the prediction of each task, especially on small data sets. The code can be downloaded at: \url{https://github.com/KomorebiTso/MTA-UNet}.
\end{abstract}


\end{frontmatter}

% \linenumbers

\section{Introduction}\label{sec1}
Satellites are important performers of space missions and play an indispensable role in fields such as communication, remote sensing, navigation, and military reconnaissance \cite{montenbruck2002satellite}.
Increasingly complicated aerospace missions pose greater challenges to the stability and reliability of satellites \cite{kodheli2020satellite}. The components installed on satellite motherboards inevitably generate much heat during operation due to their high power density \cite{chen2020heat}. The heat then leads to a raft of thermal responses such as thermal deformation, thermal buckling, and thermally induced vibration, affecting the accuracy of the satellite mission \cite{zhengchun2016design,liu2022thermal}. Ground-based simulation test and numerical analysis are two traditional methods commonly adopted in engineering to analyze thermo-mechanical coupling problems such as thermal stress and thermal deformation of satellites \cite{stohlman2018coupled,azadi2017thermally,johnston2000thermally,shen2019thermal}. However, these two methods are usually costly and require a long computational time. As a result, in terms of optimization of the design of thermal control measures and structural layout for satellites, the iterative design by the above two methods is likely to face a significant cost and computational burden 
or even fail to be completed in a given time \cite{zhao2021surrogate,chen2018practical,cuco2015multi}. It is urgent and necessary to conduct a real-time analysis of the thermo-mechanical coupling problems of the satellite. This paper aims to achieve fast and accurate predictions of thermal stress and thermal deformation of satellite motherboards when given a temperature field.

A common method for the aforementioned engineering requirements is to build a surrogate model, which implements a high-precision prediction quickly. Once iterated in optimization procedure, surrogate model could achieve a compromise between computational accuracy and computational cost, improving the efficiency of prediction tasks \cite{chen2019satellite,cuco2015multi,yao2012surrogate}. At present, the mainstream surrogate model methods include polynomial-based response surface \cite{goel2009comparing}, support vector machine regression \cite{clarke2005analysis}, radial basis function \cite{yao2011concurrent}, and Kriging interpolation \cite{zhang2019regularization}. However, when dealing with ultra-high dimensional regression problems, these traditional surrogate modeling methods all face the challenge of “curse of dimensionality” (i.e. the difficulty in constructing surrogate models between high-dimensional variables). As artificial intelligence advances, many deep learning-based surrogate model methods have been applied to the prediction of physics-related problems, and neural networks such as fully connected neural networks \cite{zakeri2019deep}, fully convolutional neural networks \cite{edalatifar2021using}, and conditional generative adversarial networks \cite{farimani2017deep} have been widely used. In particular, the U-Net, as an advanced neural network, is coming into vogue. Rishi Sharma et al. \cite{sharma2018weakly} take different boundary conditions as input and applies the U-Net to predict the temperature field in a two-dimensional plane. Cheng et al. \cite{chen2019u} analyze the two-dimensional velocity and pressure fields around arbitrary shapes in laminar flow through the U-Net. In addition, Saurabh Deshpande et al. \cite{deshpande2021fem} use the U-Net to predict the large deformation response of hyperelastic bodies under loading. These related works prove the potential of U-Net in ultra-high dimensional regression problems.

The focus of the aforementioned deep learning-based surrogate models is to dig a huge number of training data built by finite element or experiments. Nevertheless, these models are driven by data loss only, and they often suffer from high costs of training data or a small number of training data. Physics-informed neural network (PINN) \cite{raissi2019physics} is an effective way to alleviate this problem. In the face of the physics prediction tasks where some existing experience has been accumulated in the real world, the PINN method enables the integration of prior physical information with neural networks, so as to reduce the dependence of neural networks on large amounts of training data. Raissi et al. \cite{raissi2019physics} combine the residual error of PDEs with the loss function of FCN and predict the velocity and pressure fields around a cylinder, while Liu et al. \cite{liu2022temperature} use PINN to solve the forward and inverse problem of the temperature field. In addition, motivated by PINN, PDE-based loss function can also be explored for surrogate modeling. Bao et al. \cite{bao2022physics} and Zhao et al. \cite{zhao2021physics} discrete the heat conduction equation by finite difference, making it a loss function. 

It is noticed that all the physics prediction models mentioned above are only trained in a Single-Task Learning (STL) manner (i.e., each task is trained separately). When faced with multiple physics prediction tasks, the STL has two drawbacks. On the one hand, the separate training for each sub task takes up much memory space and computational time. On the other hand, if each sub task is modeled separately, the STL tends to ignore the relationship between tasks, such as correlation, conflict, and constraint, making it hard to achieve high precision for the prediction of multiple tasks. 
The Multi-Task Learning (MTL) shows more obvious advantages in dealing with such multiple tasks. In the MTL, multiple tasks share one model, which reduces memory usage and improves inference speed. MTL combines related tasks together and enables the tasks to complement each other by sharing feature information between tasks, which can effectively reduce over-fitting and improve prediction performance \cite{zhang2018overview,zhang2021survey}.

Existing MTL is divided into soft parameter sharing and hard parameter sharing. In recent years, there have been a proliferation of research and successful practices in the industrial community on soft parameter sharing, such as the Cross-Stitch \cite{misra2016cross}, PLE \cite{tang2020progressive}, ESMM \cite{ma2018entire} and SNR \cite{ma2019snr}. In particular, the MMOE \cite{ma2018modeling} proposed by Google has gained great success. It consists of expert networks shared by multiple tasks and task-specific gate networks which can decide what to share. Such a flexible design allows the model to automatically learn how to assign expert parameters based on the relationship of the low-level tasks. The gate networks thus effectively improve the generalization performance of the model and the prediction accuracy across tasks. The MTL has made remarkable achievements in the field of computer vision and has been widely applied in engineering problems such as recommendation systems. However, such MTL models have rarely been used in high-dimensional regression problems with multiple physics tasks.     

In this paper, the regression tasks of temperature fields to thermal deformation and thermal stress of satellite motherboards are regarded as mappings between images. To achieve accurate and fast prediction of these multiple physics tasks, three strategies are adopted. 
\begin{enumerate}
\item [(1)] This paper integrates the advantages of the U-Net and MTL, and proposes the Multi-Task Attention UNet network (MTA-UNet). This network not only shares feature information between high-level layers and low-level layers, but also shares feature information between different tasks. Specially, it shares the parameters in a targeted way through the attention mechanism. The MTA-UNet effectively reduces the training time of the model and improves the accuracy of prediction compared with STL U-Net.  
\item [(2)] A Physics-informed approach is applied in training the deep learning-based surrogate model, where the finite difference is applied to discrete thermoelastic and thermal equilibrium equations. The equations are encoded into a loss function to fully exploit the existing physics knowledge. 
\item [(3)] Faced with multiple physics tasks, an uncertainty-based loss balancing strategy is adopted to weigh the loss functions of different tasks during the training process. This strategy solves the problem that the training speed and accuracy between different tasks are difficult to balance if trained together, effectively reducing the phenomenon of competition between tasks.
\end{enumerate}

The rest of the paper is structured as follows. Section \ref{sec2} presents the specific definition of the mathematical model. Section \ref{sec3} introduces the strategies applied in the construction of deep learning-based MTA-UNet in detail. Section \ref{sec4} shows the training steps and the experimental results. Section \ref{sec5} is the conclusion.

\section{Mathematical Modeling of Thermal Stress and Deformation Prediction}\label{sec2}
In this paper, as previously defined in \cite{chen2020heat} and \cite{chen2018practical}, a two-dimensional satellite motherboard with partial openings is used as a study case. The objective is to realize a rapid prediction of thermal stress and thermal deformation. 

\begin{figure*}[htb]
	\centering
	{\includegraphics[scale=0.35]{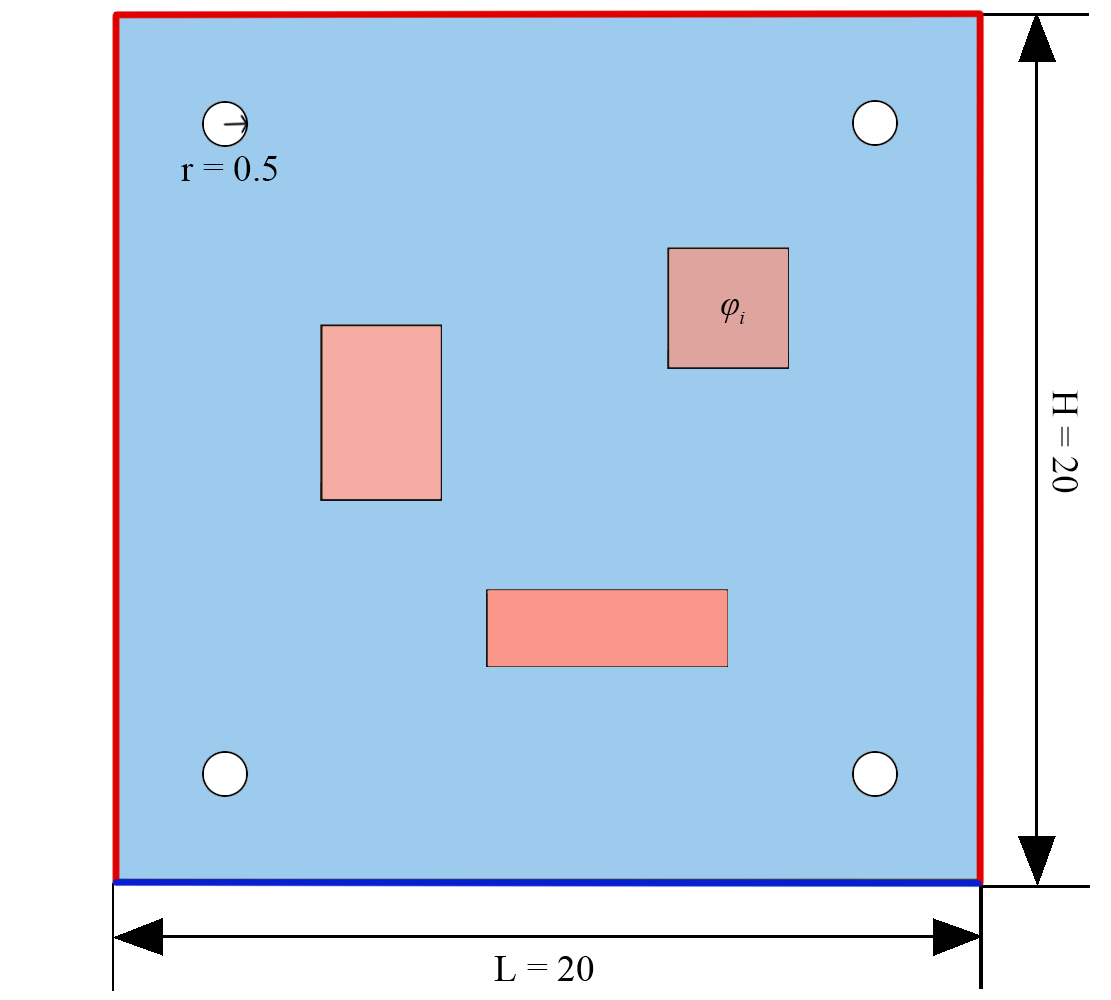}}
	\caption{Schematic diagram of a two-dimensional satellite motherboard.}\label{motherboard}
\end{figure*}

The satellite motherboard with four circular holes is shown in Fig.\ref{motherboard}, and the holes represent screw holes in engineering assembly. There are $n$ electronic components installed on the motherboard, which generate a large amount of heat during operation and can be regarded as internal heat sources. It is assumed that the temperature of the satellite motherboard changes $T$ under the joint action of the internal heat sources and external environment. Due to the thermal expansion and contraction, the elastomer tends to undergo thermal deformation, and the deformation trend is restricted to a certain extent because of the external constraints and mutual constraints between various parts of the body. With thermal stress generated inside the elastomer, the thermal stress in turn leads to new additional tension and affects the thermal deformation. 

In this case, non-displacement boundary conditions are adopted at the edges of the holes. The thermoelastic properties on the motherboard tend to be isotropic, with the parameters consistent with the thermoelastic properties of aluminum. The linear elasticity coefficient of the motherboard is considered to be around the reference temperature $T_{0}=273$ K and is assumed not to change with temperature. Non-stress and adiabatic boundary conditions are adopted on the outer boundary of the motherboard. The side length of the rectangular computational domain is set to $L=H=20$ cm. The radius of the holes is $r=0.5$ cm, and the thermal conductivity coefficient within the domain is $k = 1$ W/(m $\times$ k). The coefficient of linear expansion is $\alpha = 1e-5 $, Young’s modulus is $E = 50e3$, and Poisson’s ratio is $\mu = 0.2$.

The thermoelasticity motherboard is a two-dimensional planar elastomer and its thermal stress components and temperature satisfy the following equations:
\begin{equation}\label{1}
	\left\{\begin{array}{l}
\sigma_{xx}=\frac{E}{1-\mu^{2}}\left(\frac{\partial u_{x}}{\partial x}+\mu \frac{\partial u_{y}}{\partial y}\right)-\frac{E \alpha T}{1-\mu} \\
\sigma_{yy}=\frac{E}{1-\mu^{2}}\left(\frac{\partial u_{y}}{\partial y}+\mu \frac{\partial u_{x}}{\partial x}\right)-\frac{E \alpha T}{1-\mu} \\
\sigma_{x y}=\frac{E}{2(1+\mu)}\left(\frac{\partial u_{y}}{\partial x}+\frac{\partial u_{x}}{\partial y}\right)
\end{array}\right.
\end{equation}
where $\sigma_{xx}$ means the x-directional thermal stress, $\sigma_{yy}$ means the y-directional thermal stress, and $\sigma_{xy}$ denotes the tangential directional thermal stress. In addition, $u_{x}$ and $u_{y}$ represent x-directional thermal displacement and y-directional thermal displacement respectively. According to the differential equation of thermal equilibrium, the $u_{x}$ and $u_{y}$ satisfy the following set of equations:
\begin{equation}\label{2}
\left\{\begin{array}{l}
\frac{\partial^{2} u_{x}}{\partial x^{2}}+\frac{1-\mu}{2} \frac{\partial^{2} u_{x}}{\partial y^{2}}+\frac{1+\mu}{2} \frac{\partial^{2} u_{y}}{\partial x \partial y}-(1+\mu) \alpha \frac{\partial T}{\partial x}=0 \\
\frac{\partial^{2} u_{y}}{\partial y^{2}}+\frac{1-\mu}{2} \frac{\partial^{2} u_{y}}{\partial x^{2}}+\frac{1+\mu}{2} \frac{\partial^{2} u_{x}}{\partial x \partial y}-(1+\mu) \alpha \frac{\partial T}{\partial y}=0
\end{array}\right.
\end{equation}
and the boundary conditions further satisfy: 
\begin{equation}\label{3}
\left\{\begin{array}{l}
l\left(\frac{\partial u_{x}}{\partial x}+\mu \frac{\partial u_{y}}{\partial y}\right)_{s}+m \frac{1-\mu}{2}\left(\frac{\partial u_{x}}{\partial y}+\frac{\partial u_{y}}{\partial x}\right)_{s}=l(1+\mu) \alpha(T)_{s} \\
m\left(\frac{\partial u_{y}}{\partial y}+\mu \frac{\partial u_{x}}{\partial x}\right)_{s}+l \frac{1-\mu}{2}\left(\frac{\partial u_{y}}{\partial x}+\frac{\partial u_{x}}{\partial y}\right)_{s}=m(1+\mu) \alpha(T)_{s}
\end{array}\right.
\end{equation}
Assuming that the direction of the outer normal to the elastic plane is $\textbf{N}$, then $l$ and $m$ can be expressed as the directional cosine of the $x$ axis and $y$ axis, respectively.
%that is $l=\cos (\textbf{N}, \textbf{x}), m=\cos (\textbf{N}, \textbf{y})$.

Based on the above set of equations and the corresponding parameters, the thermal stress components and the thermal displacement components can be calculated when given an arbitrary temperature field $T$. However, the numerical methods are usually costly and time-consuming. In this paper, a deep learning-based surrogate model is constructed to quickly predict five physics components of satellite motherboard, $u_{x}$, $u_{y}$, $\sigma_{xx}$, $\sigma_{yy}$, and $\sigma_{xy}$. The five multiple physics prediction tasks are divided into three multiple sub tasks, $u_{x}$ and $\sigma_{xx}$ in task 1, $u_{y}$ and $\sigma_{yy}$ in task 2, and $\sigma_{xy}$ in task 3. Three sub tasks are trained together and the temperature field $T$ is the input of model.

\section{Method}\label{sec3}
\begin{figure*}[htb]
	\centering
	{\includegraphics[scale=0.13]{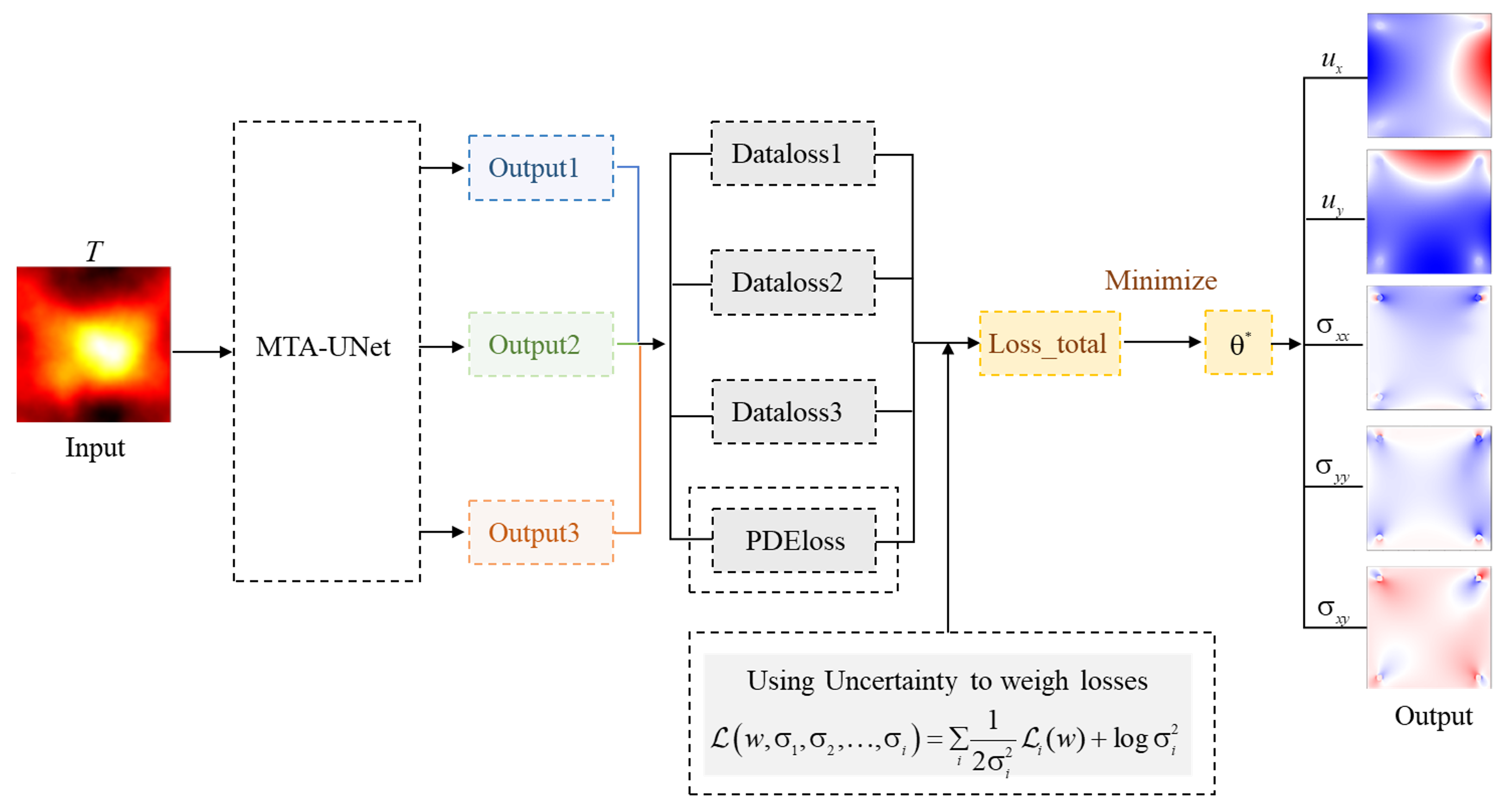}}
	\caption{The frame diagram of the main technical strategy. This paper applies three strategies when build the surrogate model, the MTA-UNet network, physics-informed model, and uncertainty-based loss balancing method respectively.}\label{framework}
\end{figure*}

This section details the technical strategies proposed during the construction of the deep learning surrogate model. Firstly, on account of multiple physics tasks, to reduce the training cost and improve the prediction performance of the model, we design a neural network structure MTA-UNet in section \ref{sec31}. Subsequently, in section \ref{sec32}, a physics-informed approach is applied to reduce the sample size of training data. Lastly, section \ref{sec33} adopts a uncertainty-based loss balancing strategy to weight the loss function of multiple tasks, adjusting the training speed and accuracy of multiple tasks. Fig.\ref{framework} shows the frame diagram of the main technical strategies.

\subsection{MTA-UNet Network Structure}\label{sec31}
\begin{figure*}[htb]
	\centering
	{\includegraphics[scale=0.16]{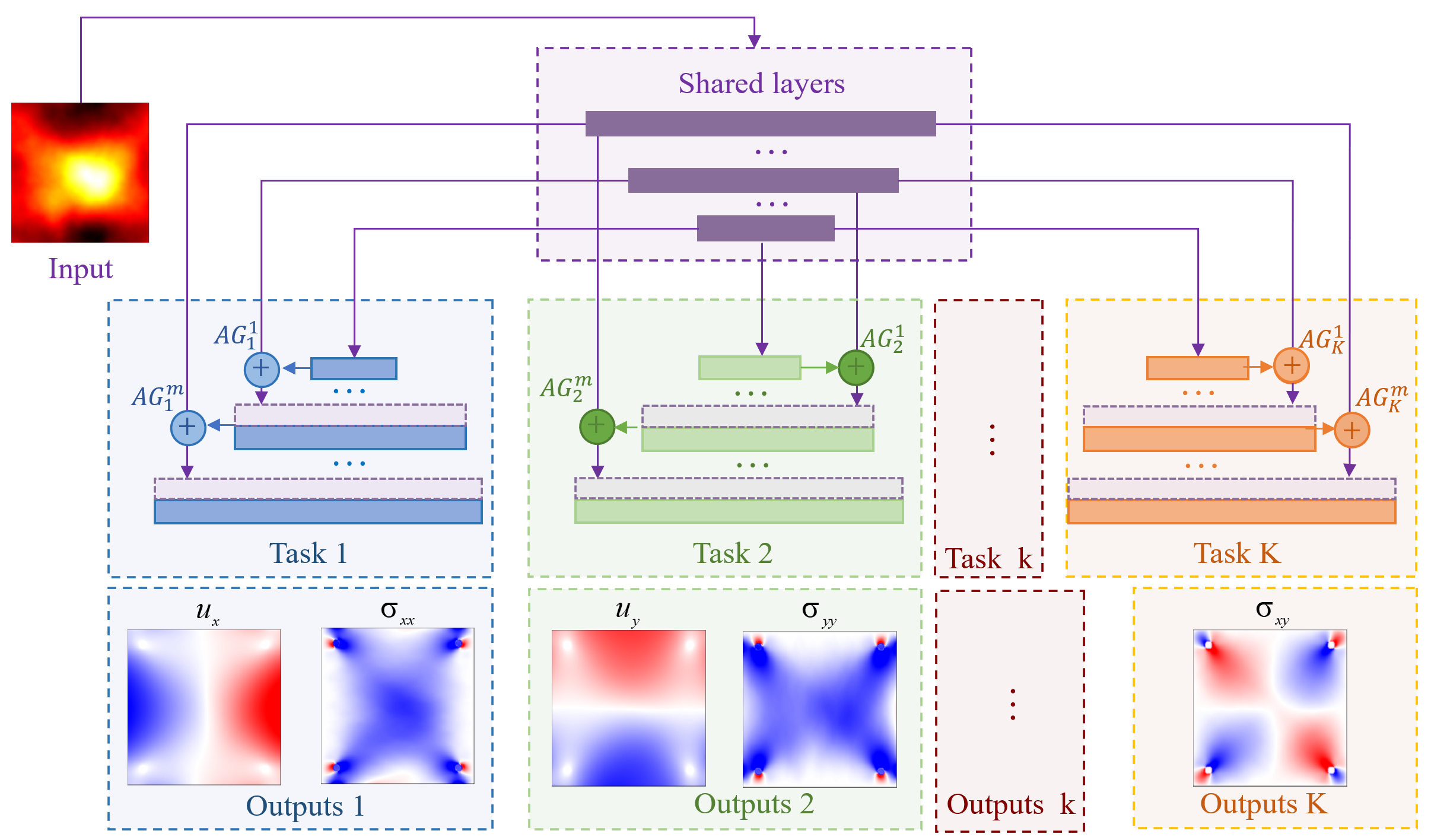}}
	\caption{Structure of MTA-UNet.}\label{MTA-UNet}
\end{figure*}

This section describes the MTA-UNet in detail, and Fig.\ref{MTA-UNet} shows the actual structure. This model takes the U-Net as the basic architecture, and the improvement is the sharing method between different tasks. In MTA-UNet, multiple sub tasks are trained together and can share feature parameters with other tasks. Different tasks share the coding layer and each of them has their specific decoding layer. In addition, when the features of different layers are concatenated, feature selection is carried out through the Attention Gate (AG). The AG distinguishes which shared features are highly correlated with specific tasks and extracts the relevant shared features from the shared coding layer. The MTA-UNet combines the merits of U-Net and MTL, having the advantage of a “purposeful” combination of feature parameters from different layers and physics tasks. This design fully learns information from the training sets, reducing the training time of the model and memory space of the system and also improving the prediction performance of the model. This section is divided into the architecture of the MTA-UNet (\ref{sec311}), the sharing mechanism of the MTA-UNet (\ref{sec312}), and the Attention Gate of the MTA-UNet (\ref{sec313}).

\subsubsection{Architecture of the MTA-UNet Network}\label{sec311}
\begin{figure*}[htb]
	\centering
	{\includegraphics[scale=0.1]{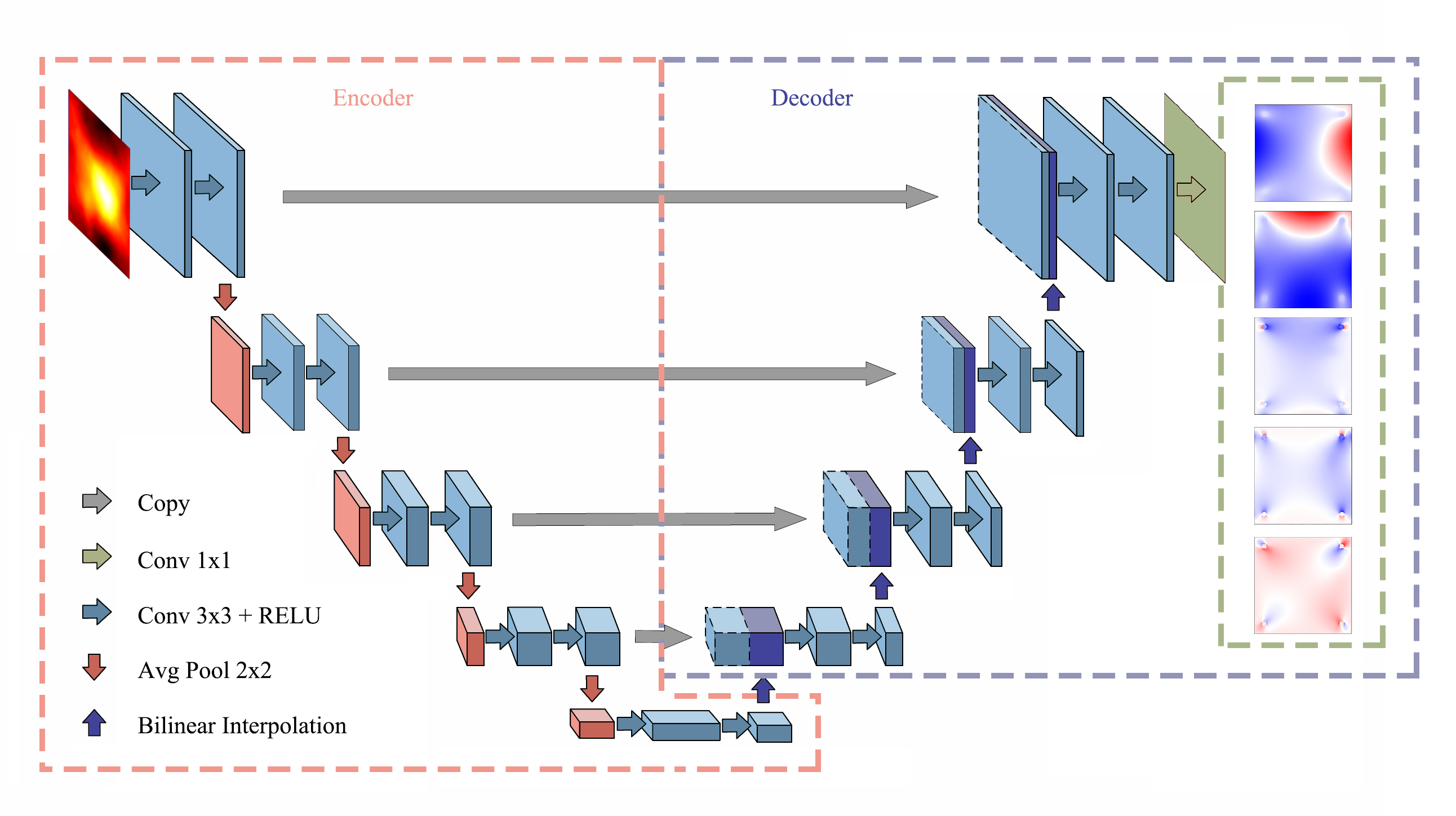}}
	\caption{Structure of U-Net.}\label{UNet}
\end{figure*}

%To fully mine information from the training sets,
To fully share feature information of different layers, MTA-UNet adopts the U-Net \cite{ronneberger2015u} as its basic architecture. In recent years, the U-Net has exhibited good performance when dealing with various ultra-high dimensional regression problems. According to Fig.\ref{UNet}, U-Net has a U-shaped symmetric structure which combines feature parameters of different layers. Owing to this characteristic, the U-Net network is suitable for image-to-image regression tasks as it takes multi-scale feature fusion into account and increases the amount of information.

The MTA-UNet network built in this paper fully retains the advantages of the U-Net network. The first half of MTA-UNet is a classical downsampling process, which is the feature extraction part similar to the coding process, capable of learning multi-scale features of an image through pooling and convolution operations. The second half is an upsampling process similar to the decoding part, capable of restoring the size of the image layer by layer through deconvolution operations. These feature maps are effectively used in subsequent calculations through skip connections. 

In MTA-UNet, the downsampling layers are shared layers and the upsampling layers are exclusive to multiple tasks, respectively. The MTA-UNet consists of two repetitive applications of 3x3 convolutions (padding=1), and each convolution is followed by a BatchNorm2d normalized activation function layer and a 2x2 max-pooling operation with a stride of 2. We double the number of feature channels in each step of the shared downsampling and, in contrast, halve the number of feature channels in each step of the exclusive upsampling. After the feature selection through the AG, feature parameters of upsampling are skip connected to downsampling with the corresponding scale of the feature map. In the last layer, each component feature vector is mapped to its corresponding class by a 1x1 convolution.

\subsubsection{Feature Sharing Mechanism of MTA-UNet}\label{sec312}
To enable the sharing of feature parameters across multiple physics tasks, the MTA-UNet uses a partially shared coding layer to enable simultaneous learning of multiple tasks. During the downsampling process, multiple tasks share the coding layer, each having an independent decoding layer. This information concatenation operation of shared and specific feature parameters realizes the deep sharing of multi-task feature parameters. 

For the task $k$ in the given $K$ tasks, $k=1,2, \ldots, K$. The model consists of the shared downsampling function $f$, the exclusive upsampling $g_{k}$ , and $k$ task networks $h_{k}$. The shared layer follows the input layer, and the task networks are built upon the output of the shared-bottom. The $y_{k}$ is the output of the task $k$, follows the corresponding task-specific tower. Assuming that for the task $k$, the $m^{th}$ downsampling layer of the model is $f^{m}$,  and its corresponding upsampling layer is $g_{k}^{m}$, the output of the task $k$ is:
\begin{equation}
	\begin{aligned}\label{4}
		y_{k}=h_{k}\left(\text {SkipConnect}\left(AG_{k}^{m}\left(f^{m}, g_{k}^{m}\right), g_{k}^{m}\right)\right)
	\end{aligned}
\end{equation}
where $AG_{k}^{m}$ denotes the AG operation for the $m^{th}$ layer of the task $k$.

\subsubsection{Attention Gate of MTA-UNet}\label{sec313}
Inspired by CNN-based MTL models such as MMOE \cite{ma2018modeling} and PLE \cite{tang2020progressive}, the AG operation is added to the MTA-UNet model to realize the selection of shared features. In this way, shared features that are highly relevant to specific tasks are extracted from the shared network layer.

\begin{figure*}[htb]
	\centering
	{\includegraphics[scale=0.11]{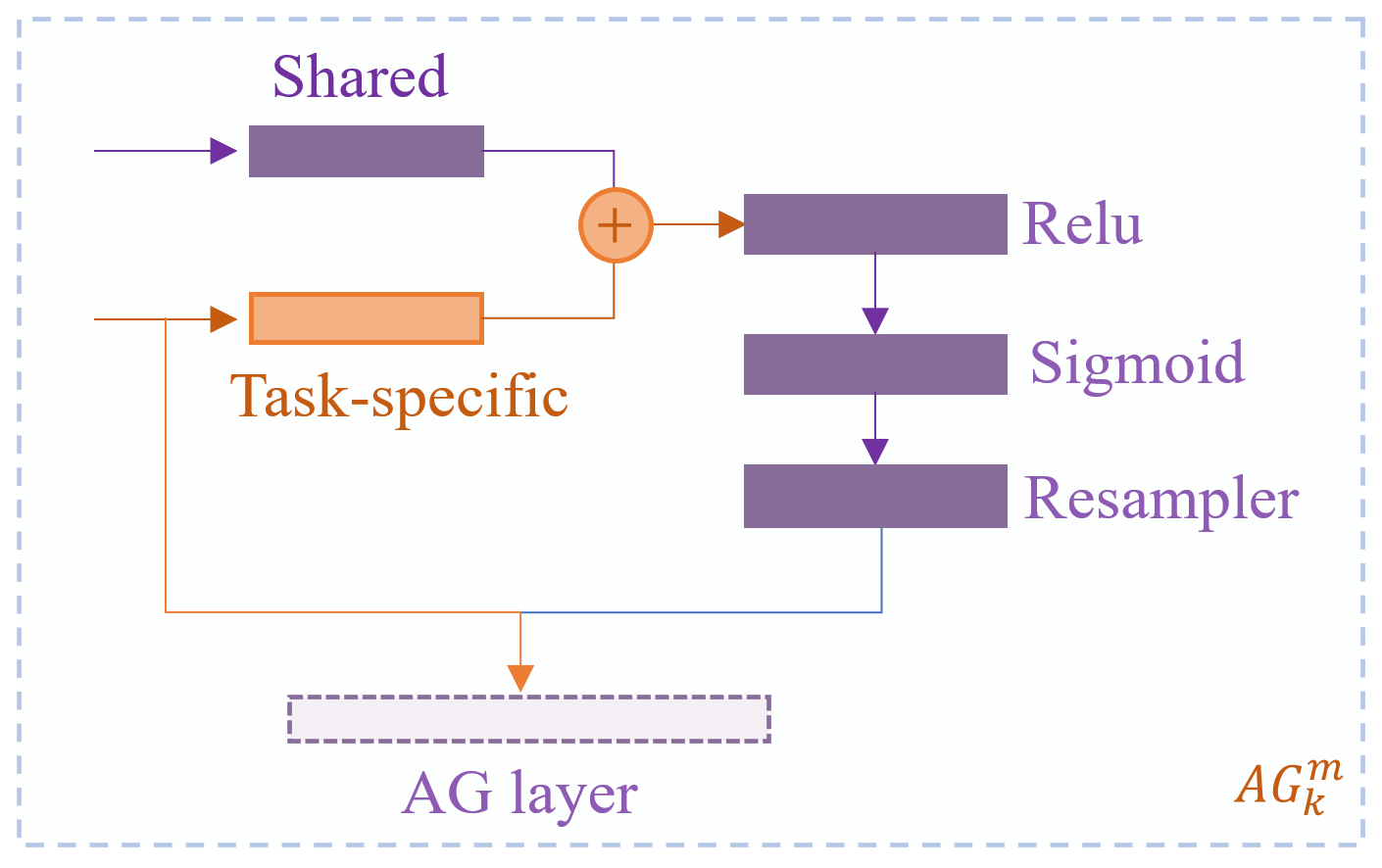}}
	\caption{Attention Gate operation. }\label{attention}
\end{figure*}

Fig.\ref{attention} illustrates the AG operation of task $k$, where a series of operations such as Relu, Sigmoid, and resampler are performed to evaluate the similarity between $f^{m}$ and $g_{k}^{m}$. The $m^{th}$ upsampling layer $f^{m}$ and its corresponding downsampling layer $g_{k}^{m}$ are input of $AG_{k}^{m}$. Since the underlay network $g_{k}^{m}$ is generally more accurate, when the similarity between $f^{m}$ and $g_{k}^{m}$ is calculated, the features in $f^{m}$ with higher similarity with $g_{k}^{m}$ are given higher weights, equals chosen of feature parameters of $f^{m}$. The selection is reflected in the calculation of the weight coefficient, and the operation is presented as:
\begin{equation}
	\begin{aligned}\label{5}
		A G_{k}^{m}\left(f^{m}, g_{k}^{m}\right)=\sum_{i=1}^{L_{x}} \operatorname{Similarity}\left(f^{m}, g_{k}^{m}\right) \times f^{m}
	\end{aligned}
\end{equation}

\subsection{A Physics-informed Training Strategy}\label{sec32}
To make full use of the prior physical laws among the physical quantities being predicted, reducing the size of the training sample, a physics-informed strategy is used in the training process. The existing physics knowledge (Eq.\ref{1}-Eq.\ref{3}) is discretized by the Finite Difference Method (FDM) and integrated into a loss function to construct a physics-informed surrogate model. The FDM is a numerical solution method that expresses basic equations and boundary conditions in the form of function approximation.

% The finite difference method (FDM) is an approach to solving differential equations that are difficult to solve or from which it is impossible to obtain an analytical solution. The so-called difference method is to represent the approximation of basic equations and boundary conditions with difference equations and replace the problem of solving differential equations with the problem of solving algebraic equations.

\begin{figure*}[htb]
	\centering
	{\includegraphics[scale=0.3]{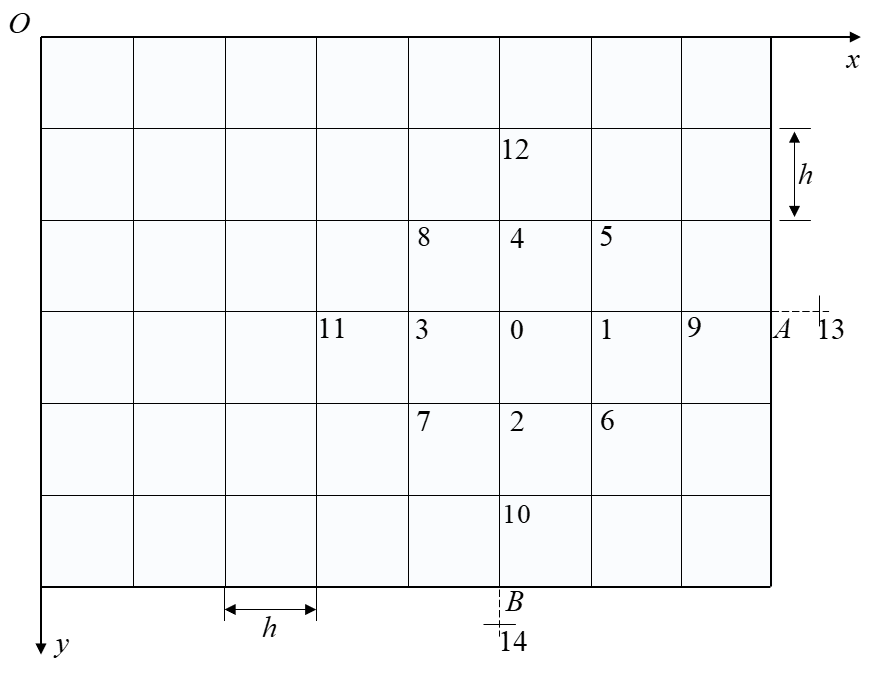}}
	\caption{Grid diagram of finite differences.}\label{FDM}
\end{figure*}

It was assumed that an arbitrary two-dimensional elastomer is divided into a uniform grid as shown in Fig.\ref{FDM}, in which the intersections of the lines are the nodes.

% Let $f=f(x,y)$ be a continuous function which the elastomer satisfies, and it can represent a stress component, a displacement component, temperature, etc. Assuming that the grid spacing is sufficiently small, then $\left|x-x_{0}\right|=h$, and at node 0, the central difference formula for the first and second derivatives with respect to $x$ at node 0 can be obtained:

Let $f=f(x,y)$ be a continuous physical quantity, which can be expressed as stress component, displacement component, temperature, etc. Assuming that the grid spacing is sufficiently small, then $\left|x-x_{0}\right|=h$. The central difference formulas of the first and second derivatives of $x$ at node 0 can be obtained :

\begin{equation}
	\begin{aligned}\label{6}
\left(\frac{\partial f}{\partial x}\right)_{0} \approx \frac{f_{1}-f_{3}}{2 h},\left(\frac{\partial^{2} f}{\partial x^{2}}\right)_{0} \approx \frac{f_{1}+f_{3}-2 f_{0}}{h^{2}}
	\end{aligned}
\end{equation}
Similarly, the central difference formulas for the first and second derivatives at node 0 in the y direction are:
\begin{equation}
	\begin{aligned}\label{7}
\left(\frac{\partial f}{\partial y}\right)_{0} \approx \frac{f_{2}-f_{4}}{2 h},\left(\frac{\partial^{2} f}{\partial y^{2}}\right)_{0} \approx \frac{f_{2}+f_{4}-2 f_{0}}{h^{2}}
	\end{aligned}
\end{equation}
According to the above equations, the central difference formulas for mixed second derivatives are: 
\begin{equation}
	\begin{aligned}\label{8}
\left(\frac{\partial^{2} f}{\partial x \partial y}\right)_{0}=\left[\frac{\partial}{\partial x}\left(\frac{\partial f}{\partial y}\right)\right]_{0}=\frac{\left(\frac{\partial f}{\partial y}\right)_{1}-\left(\frac{\partial f}{\partial y}\right)_{3}}{2 h} \approx \frac{1}{4 h^{2}}\left[\left(f_{6}+f_{8}\right)-\left(f_{5}+f_{7}\right)\right]
	\end{aligned}
\end{equation}
For boundary points where central difference cannot be used, forward difference or backward difference are used for discretization:
\begin{equation}
	\begin{aligned}\label{9}
	\left(\frac{\partial f}{\partial x}\right)_{0} \approx \frac{-3 f_{0}+4 f_{1}-f_{9}}{2 h} \approx \frac{3 f_{0}-4 f_{3}+f_{11}}{2 h},
	\end{aligned}
\end{equation}

\begin{equation}
	\begin{aligned}\label{10}
	\left(\frac{\partial f}{\partial y}\right)_{0} \approx \frac{-3 f_{0}+4 f_{2}-f_{10}}{2 h} \approx \frac{3 f_{0}-4 f_{4}+f_{12}}{2 h}
	\end{aligned}
\end{equation}

To deal with the prediction tasks of thermal stress and thermal displacement, a neural network $\hat{Y}(x, \theta)$ is constructed, where $\theta=\left\{W^{\ell}, b^{\ell}\right\}_{1 \leq \ell \leq D}$, and $W^{\ell}$ is the weight matrix, $b^{\ell}$ is the deviation vector in the neural network, and the physics-informed loss function of the neural network is defined as: 
\begin{equation}
	\begin{aligned}\label{11}
\mathcal{L}_{\text {PINN}}(\theta)=\mathcal{L}_{\text {data }}+\mathcal{L}_{\text {pde}}
	\end{aligned}
\end{equation}
where $\mathcal{L}_{\text {data}}$ is the data loss obtained from labeled data with the loss function of Mean Square Error (MSE), and the data loss of the task $k$ is defined as: 
\begin{equation}
	\begin{aligned}\label{12}
\mathcal{L}_{\text {data}_{k}}=\frac{1}{\left|\mathcal{P}_{f}\right|} \sum_{(\mathbf{x}_{k}, \mathbf{y}_{k}) \in \mathcal{P}_{f}}\left(Y_{{k}}-\hat{Y}_{{k}}\right)^{2}
	\end{aligned}
\end{equation}
where $\mathcal{P}_{f}$ denotes the size of data set, $\hat{Y}_{k}$ represents the prediction of the model, $Y_{k}$ is the ground truth.  

The aforementioned finite difference equations (Eq.\ref{6}-Eq.\ref{10}) are used to discretize the thermoelasticity and the thermal equilibrium equations, with the central difference method for interior points and a forward/backward difference method for boundary points. The discretized PDEs are encoded into a loss function $\mathcal{L}_{\text {pde}}$ ($\mathcal{L}_{\text {pde}}$ denotes the physical losses of $K$ tasks), which is defined as:
\begin{equation}
	\begin{aligned}\label{13}
\mathcal{L}_{pde}=\frac{1}{\left|\mathcal{P}_{f}\right|} \sum_{(x, y) \in \mathcal{P}_{f}}\left(\mathcal{L}_{\sigma_{xx_{-}} pde}+\mathcal{L}_{\sigma_{yy_{-}}pde}+\mathcal{L}_{\sigma_{xy_{-}}pde}+\mathcal{L}_{u_{x_{-}}p d e}+\mathcal{L}_{u_{y_{-}}p d e}\right)
	\end{aligned}
\end{equation}

Where $\mathcal{L}_{u_{x_{-}}pde}$, $\mathcal{L}_{u_{y_{-}}pde}$, $\mathcal{L}_{\sigma_{xx_{-}} pde}$, $\mathcal{L}_{\sigma_{xy_{-}} pde}$, $\mathcal{L}_{\sigma_{yy_{-}} pde}$ mean the PDE loss of Eq.\ref{1} and Eq.\ref{2} respectively.The boundary loss $\mathcal{L}_{bc}$ is added to the physical loss function matrix $\mathcal{L}_{pde}$ in the form of the matrix mask. A convolution operation can also realize the physics-informed strategy.

By training the neural network and minimizing the loss function until the best parameter $\theta^{*}$ is found, we construct a physics-informed surrogate model of the thermal deformation and thermal stress of satellite motherboards.

\subsection{Uncertainty-based Multi-Task Loss Balancing Strategy}\label{sec33}
The training goal of the MTA-UNet model is to minimize the difference between the prediction results and the labels. For MTL, however, errors between multiple tasks features cause competition between multiple tasks. To overcome this seesaw phenomenon, it is an integral part of MTL to balance the training speed and accuracy of multiple tasks through some loss balancing strategies.  

Since the loss functions of different tasks tend to have different magnitudes, this paper first normalized their inputs. Subsequently, 
to solve the problem of competition between
different tasks, the most direct approach is to use the weight coefficient $w_{i}$ to adjust the proportions of the loss functions of different tasks, as shown in the following equation.

\begin{equation}
	\begin{aligned}\label{13}
\operatorname{Loss}=\sum_{k}^{K} w_{k} \times \operatorname{Loss}_{k}
	\end{aligned}
\end{equation}

However, this loss balancing method of fixed weights may not be applicable in some cases as different tasks are very sensitive to the setting of $w_{i}$, and different settings of $w_{i}$ vary greatly in terms of performance. 

We would like to use a dynamic weight method to balance the loss functions of different tasks. The mainstream methods include GradNorm \cite{chen2018gradnorm}, DWA \cite{liu2019end}, DTP \cite{guo2018dynamic}, and Uncertainty \cite{kendall2018multi,xiang2021self}, among which GradNorm consumes a longer computational time as it requires gradient computation, while DWA and DTP need additional weighting operations. This paper adopts the uncertainty-based method to dynamically adjust the weight coefficients for the loss functions of different tasks. The model in this study is constructed based on task-dependent and homoscedastic uncertainty in aleatoric uncertainty \cite{kendall2018multi}.

For a certain task $k$, the model predicts both a prediction $y_{k}$ and the homoscedastic uncertainty $\sigma_{k}$ of the model. Thus the loss function for MTL is defined as: 
\begin{equation}
	\begin{aligned}\label{14}
\mathcal{L}\left(W, \sigma_{1}, \sigma_{2}, \ldots, \sigma_{k}\right)=\sum_{k} \frac{1}{2 \sigma_{k}^{2}} \mathcal{L}_{k}(W)+\log \sigma_{k}^{2}
	\end{aligned}
\end{equation}

where $W$ means the weight matrix, $\mathcal{L}_{k}(W)$ is the loss function of task k. $\sigma_{k}$ means noise of the model of task $k$, $\frac{1}{2 \sigma_{k}^{2}}$ is the weight coefficient of the corresponding task, and $\text{log}\sigma_{k}^{2}$ is the regularization item to prevent $\sigma_{k}$ from learning too large. In this way, higher weights can be given to simple tasks, and the learning of other tasks are driven by them. 

For the total of $K$ regression tasks driven by physics, the overall loss function is defined as:

\begin{equation}
	\begin{aligned}\label{15}
\mathcal{L}=\sum_{k=1}^{K} \left(\frac{1}{2 \sigma_{k}^{2}} \mathcal{L}_{\text {data}_{k}}(W)+\log \sigma_{k}^{2}\right) + \frac{1}{2 \sigma_{pde}^{2}} \mathcal{L}_{pde}(W)+\log \sigma_{pde}^{2}
	\end{aligned}
\end{equation}

\section{Experiments}\label{sec4}
In this section, the training details and the results of experiments are shown. Section \ref{sec41} introduces the data set and training metrics. Section \ref{sec42} discusses the effect of dynamic loss balancing strategy, MTA-UNet, and the physics-informed method in detail, respectively.

\subsection{Training Steps}\label{sec41}
\begin{figure*}[htb]
	\centering
	{\includegraphics[scale=0.4]{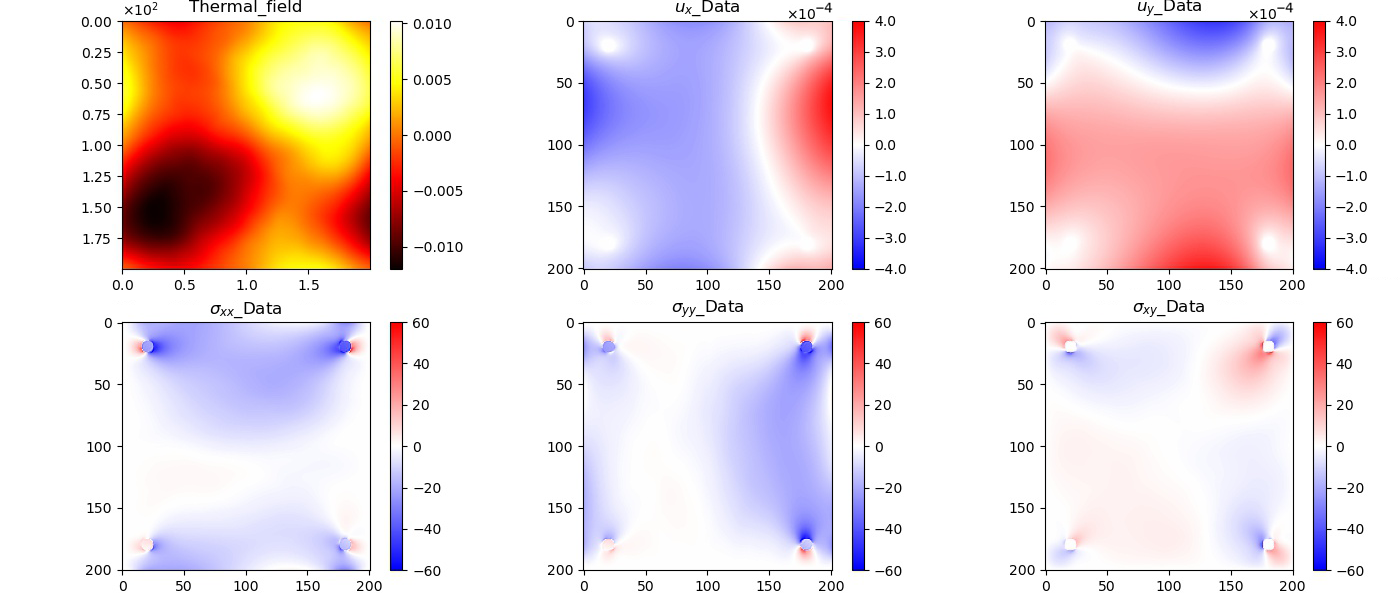}}
	\caption{A sample of data set. This figure shows a sample of the data set. The temperature field is the input of surrogate model, and $u_{x}$, $u_{y}$, $\sigma_{xx}$, $\sigma_{yy}$, $\sigma_{xy}$ are outputs.}
	\label{dataset}
\end{figure*}

\pmb{Data set:} In MTA-UNet, the input is a two-dimensional planar temperature field on a $200\times200$ uniform grid and the output is five corresponding thermal stress matrices and thermal deformation matrices, including x-directional displacement $u_{x}$, y-directional displacement $u_{y}$, x-directional thermal stress $\sigma_{xx}$, y-directional thermal stress $\sigma_{yy}$, and tangential thermal stress $\sigma_{xy}$. The OpenFOAM software is applied to build the data sets, and the computational domain is a uniform grid. Fig.\ref{dataset} shows one of the data set samples. 

The input temperature field is a Gaussian Random Field (GRF) \cite{haran2011gaussian}, and the roughness of the temperature field matrix changes when the mean and covariance of GRF are adjusted. This paper generates training sets with sample sizes of 100, 200, 500,1000, 2000, 5000, and 10,000. Firstly, the prediction of the STL model U-Net and the MTL model MTA-UNet are compared in the training set with a sufficient number of 10,000 samples to confirm the superiority of our model. Subsequently, the MTA-UNet model is trained in training sets with sample sizes of 100, 200, 500, 1000, 2000, and 5000 respectively to verify the effectiveness of the physics-informed approach. The trained models are tested in a general test set with a sample size of 500.

\pmb{Metrics:} AdamDelda \cite{zeiler2012adadelta} is selected as the optimizer and the MTA-UNet model is implemented by PyTorch 1.8. The training batch size is set to 16. The prediction performance of different models for task $k$ is measured by metrics MAE and MRE, defined as: 
\begin{equation}
	\begin{aligned}\label{15}
MAE_{k}=\frac{1}{\left|\mathcal{P}_{f}\right|} \sum_{(\mathbf{x}_{k}, \mathbf{y}_{k}) \in \mathcal{P}_{f}}\left|Y_{k}-\hat{Y}_{k}\right|
	\end{aligned}
\end{equation}

% \begin{equation}
% 	\begin{aligned}\label{15}
% MSE_{k}=\frac{1}{\left|\mathcal{P}_{f}\right|} \sum_{(\mathbf{x}_{k}, \mathbf{y}_{k}) \in \mathcal{P}_{f}}\left|Y_{k}-\hat{Y}_{k}\right|^{2}
% 	\end{aligned}
% \end{equation}

\begin{equation}
	\begin{aligned}\label{15}
M R E_{k}=\frac{1}{\left|\mathcal{P}_{f}\right|} \sum_{(\mathbf{x}_{k}, \mathbf{y}_{k}) \in \mathcal{P}_{f}} \frac{\left|Y_{k}-\hat{Y}_{k}\right|}{\hat{Y}_{k}}
	\end{aligned}
\end{equation}

\subsection{Experiment Results}\label{sec42}
\subsubsection{Effectiveness of Dynamic Loss Balancing Strategies}\label{sec421}
\begin{figure}[htbp]
 \centering
 \subfigure[Fixed weighted loss balancing strategy]{
  {\includegraphics[width=0.45\linewidth]{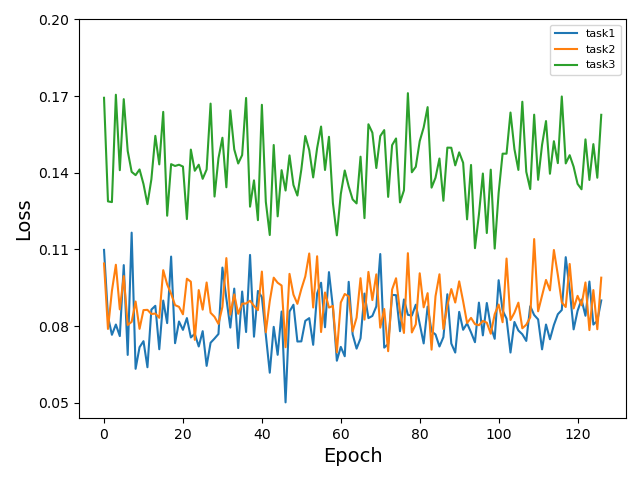}}
 }
 \quad
 \subfigure[Uncertainty-based loss balancing strategy]{
  {\includegraphics[width=0.45\linewidth]{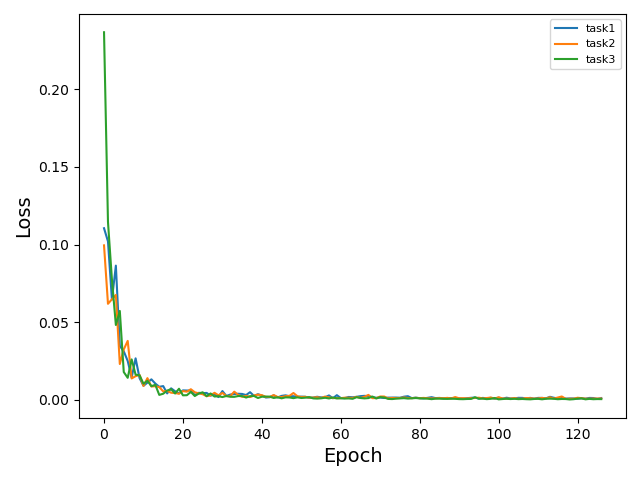}}
 }
 \caption{Training curves of different loss balancing strategies. The training curves of fixed weighted strategy shock in a certain range, making it difficult for training among multiple tasks. While the training curves of dynamic uncertainty-based loss balancing strategy decrease in a  stable speed, enable good prediction for multiple tasks simultaneously.}
 \label{balance comparastion}
\end{figure}

To verify the necessity and effectiveness of the dynamic loss balancing strategy in our MTL regression model, training is carried out with the fixed weighted loss balancing strategy and uncertainty-based dynamic loss balancing strategy, respectively. Fig.\ref{balance comparastion} shows the training curves, indicating that if the loss function of MTL is balanced by fixed weight, multiple tasks will deviate in different directions due to different task objectives. In this way, the better learning of one task decreases the accuracy of another. Unlike fixed weight, the loss functions of multiple tasks simultaneously decrease at a stable speed with the uncertainty-based dynamic balancing strategy. The dynamic strategy avoids the phenomenon of competition and enables multiple tasks to achieve good performance simultaneously.

\subsubsection{Performances of Models}\label{sec422}
We train the MTA-UNet model and the U-Net model in the training set of 10,000 samples and test them in the general test set, respectively. The MAE and MRE errors obtained from the training are shown in Table.\ref{table1} and Table.\ref{table2}. The $\sigma_{MAE}$ and $\sigma_{MRE}$ mean the standard deviations of MAE and MRE respectively, and the data in bold represent the result of MTA-UNet.

%\begin{table}[htb]
%\caption{The experimental results of different network structures}\label{table1}
%\centering
%\begin{tabular}{ccccc} 
%\hline
%\multirow{2}{*}{Option} & \multicolumn{2}{c}{$MAE$} & \multicolumn{2}{c}{$MRE/\%$}  \\ 
%\cline{2-5}
%               & U-Net      & MTA-UNet  & U-Net & MTA-UNet  \\ 
%\hline
%$u_{x}$        & 1.9754e-6 & \pmb{1.8455e-6} & 2.23 & \pmb{2.06}    \\
%$u_{x}$        & 2.0947e-6 & \pmb{1.9264e-6} & 2.39 & \pmb{2.13}    \\
%$\sigma_{xx}$  &  0.0945   & \pmb{0.0713}    & 1.27 & \pmb{0.96}    \\
%$\sigma_{xy}$  &  0.0474   & \pmb{0.0402}    & 1.36 & \pmb{1.03}    \\
%$\sigma_{yy}$  &  0.0924   & \pmb{0.0728}    & 1.29 & \pmb{0.98}    \\
%\hline
%\end{tabular}
%\end{table}

\begin{table}
\caption{The MAE of different networks}\label{table1}
\centering
\setlength{\tabcolsep}{3mm}{
\begin{tabular}{ccccc}
\toprule
\multirow{2}{*}{Task} & \multicolumn{2}{c}{$MAE$} & \multicolumn{2}{c}{$\sigma_{MAE}$} 
\\ 
\cline{2-5}
                & U-Net     & MTA-UNet     & U-Net    & MTA-UNet  \\ 
\hline
$u_{x}$         & 1.9754$\times10^{-6}$ & \bf{1.1556}\bm{$\times10^{-6}$} & $\pm$8.2727$\times10^{-7}$ & \bm{$\pm$}\bf{1.9903}\bm{$\times10^{-7}$} \\
$u_{y}$         & 2.0947$\times10^{-6}$ & \bf{1.0721}\bm{$\times10^{-6}$} & $\pm$8.3263$\times10^{-7}$ & \bm{$\pm$}\bf{2.2980}\bm{$\times10^{-7}$} \\
$\sigma_{xx}$   & 0.0945    & \bf{0.0713}    & $\pm$0.01  & \bm{$\pm$}\bf{9.5297}\bm{$\times10^{-5}$}     \\
$\sigma_{yy}$   & 0.0924    & \bf{0.0728}    & $\pm$0.01  & \bm{$\pm$}\bf{0.0098}    \\
$\sigma_{xy}$   & 0.0474    & \bf{0.0402}    & $\pm$0.01  & \bm{$\pm$}\bf{0.0087}    \\
\bottomrule
\end{tabular}}
\end{table}

\begin{table}
\caption{The MRE of different networks}\label{table2}
\centering
\setlength{\tabcolsep}{5.5mm}{
\begin{tabular}{ccccc}
\toprule
\multirow{2}{*}{Task} & \multicolumn{2}{c}{$MRE/\%$}&
\multicolumn{2}{c}{$\sigma_{MRE}$}
\\ 
\cline{2-5}
                & U-Net     & MTA-UNet     & U-Net    & MTA-UNet  \\ 
\hline
$u_{x}$         & 2.23   & \bf{1.30} & $\pm$0.0056     & \bm{$\pm$}\bf{0.0044}\\
$u_{y}$         & 2.39   & \bf{1.21} & $\pm$0.0058     & \bm{$\pm$}\bf{0.0039}\\
$\sigma_{xx}$   & 1.27   & \bf{0.96} & $\pm$0.0036     & \bm{$\pm$}\bf{0.0027}\\
$\sigma_{yy}$   & 1.29   & \bf{0.98} & $\pm$0.0039     & \bm{$\pm$}\bf{0.0027}\\
$\sigma_{xy}$   & 1.36   & \bf{1.03} & $\pm$0.0038     & \bm{$\pm$}\bf{0.0030}\\
\bottomrule
\end{tabular}}
\end{table}

\begin{figure}[htbp]
 \centering
 \subfigure[Sample 1]{
  {\includegraphics[width=0.95\linewidth]{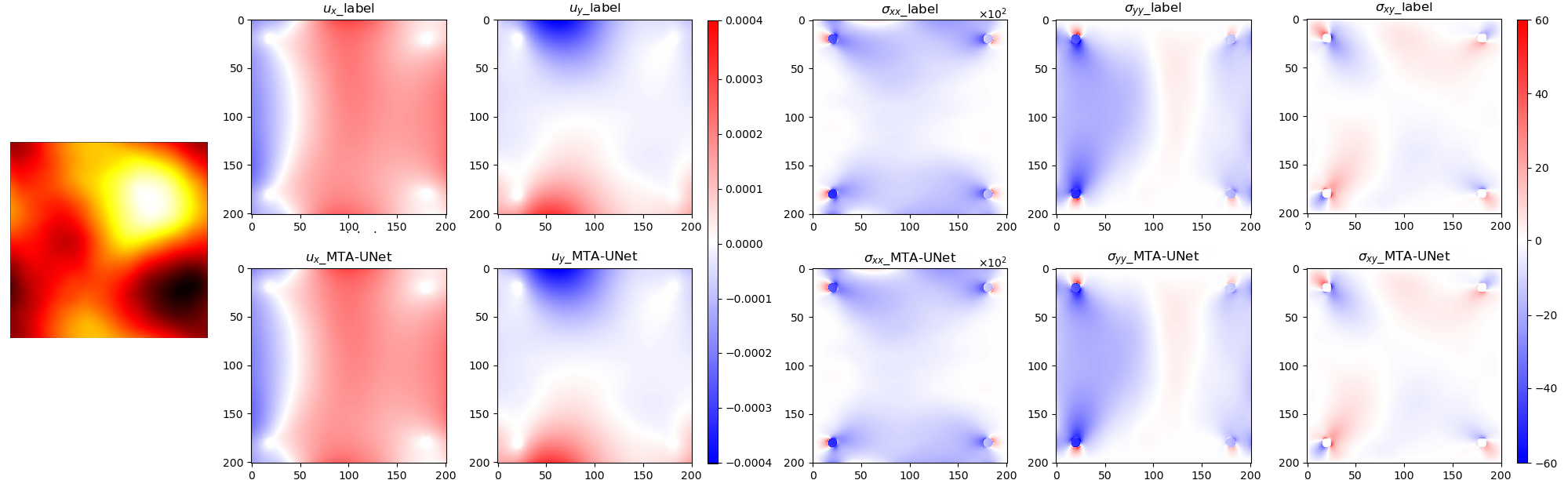}}
 }
 \quad
 \subfigure[Sample 2]{
  {\includegraphics[width=0.95\linewidth]{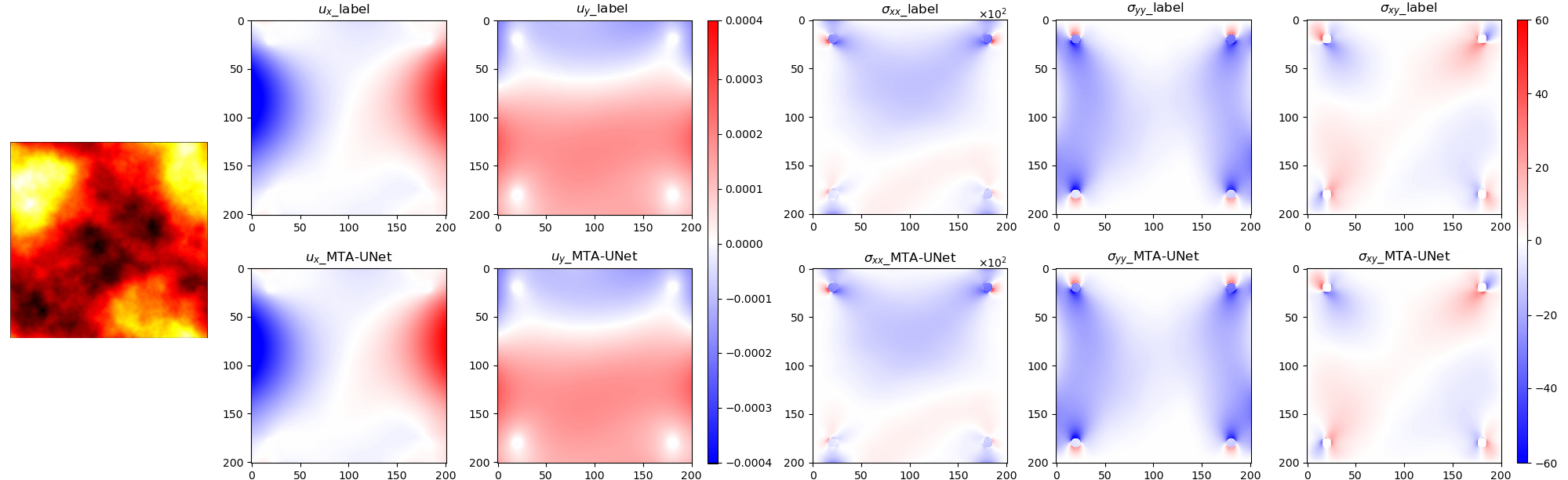}}
 }
 \caption{Prediction of the MTA-UNet model. This figure shows two samples of results by MTA-UNet. The up row of each sample is the label of $u_{x}$, $u_{y}$, $\sigma_{xx}$, $\sigma_{yy}$ and $\sigma_{xx}$, the next line are corresponding predictions of MTA-UNet.}
 \label{result1}
\end{figure}

We first investigate the prediction performance. According to Table.\ref{table2}, the minimum per-pixel MRE of the proposed MTA-UNet model is only 0.96\%, and the model shows good performance in predicting thermal stress and thermal deformation of satellite motherboards.
Fig.\ref{result1} shows the prediction result of MTA-UNet. The mean per-pixel error is calculated with high precision according to the ground truth, demonstrating the feasibility of using the MTA-UNet model for regression between ultra-high dimensional variables. In addition, as for the comparison between the STL U-Net and the MTL MTA-UNet model proposed, it can be seen that the MTA-UNet exhibits better prediction performance on each physical prediction task by fully sharing feature information between different tasks. Especially in $u_{x}$, the proposed MTL MTA-UNet model reduces the $MAE$ by at least 41\% lower than the STL U-Net model in our task. Since different tasks have different noise, the sharing of feature parameters between tasks by MTA-UNet cancels part of the noise in different directions to a certain extent, improving the prediction accuracy. According to variances of MAE and MRE for different models, the MTA-UNet model is more robust than U-Net. The sharing mechanism in MTA-UNet plays the role of data enhancement. Moreover, some parameters required in a single task are better trained by other tasks, and the sharing of feature between different tasks make sense in this respect. 

Then, the training cost has been discussed. This deep learning surrogate model can reduce the prediction time of thermal stress and thermal deformation from 2 minutes to 0.23 seconds, effectively saving the time cost, which also verifies the effectiveness of the deep learning-based thermal stress and thermal deformation surrogate model.

% \begin{figure}[htbp]
%     \centering
%     \begin{subfigure}
%     {\includegraphics[scale=0.24]{result1-1.png}}
%     \newline
%     \subcaption{Sample1}
% 	\label{result1-1} %文中引用该图片代号
% 	\end{subfigure}

%     \begin{subfigure}
%     {\includegraphics[scale=0.24]{result1-2.png}}
%     \newline
%     \centering
%     \subcaption{Sample2}
% 	\label{result1-2}
% 	\end{subfigure}
%     \caption{Prediction of the MTA-UNet model. This figure shows two samples of results by MTA-UNet. The up row of each sample is the label of $u_{x}$, $u_{y}$, $\sigma_{xx}$, $\sigma_{yy}$ and $\sigma_{xx}$, the next line are corresponding predictions of MTA-UNet.}\label{result1}
% \end{figure}

\subsubsection{Effects of the Physics-informed Strategy}\label{sec423}

To verify the contributions of the physics-informed strategy to the accuracy of the model and sample size, the MTA-UNet surrogate model is trained in data sets with different sample sizes and tested in a general test set, and the comparison of MRE between data-driven only and the physics-informed is shown in Table.\ref{table3}. The $\sigma_{MRE}$ means the standard deviation of MRE, and the data in bold represent the result of the physics-informed strategy.

% \begin{table}
% \caption{Comparison of training strategies on dataset with different scales}\label{table2}
% \centering
% \begin{tabular}{ccccccc} 
% \toprule
%     Dataset Scale   & Training Strategy & $u_{x}$ & $u_{y}$ & $\sigma_{xx}$ & $\sigma_{yy}$ & $\sigma_{xy}$ \\ 
% \hline
% \multirow{2}{*}{200} & Data      & 6.46 & 6.42 & 4.43 & 4.55 & 5.52  \\
%                     & Data\&PDE & \pmb{6.07} & \pmb{6.02} & \pmb{4.0}  & \pmb{4.17} & \pmb{5.11}  \\ 
% \hline
% \multirow{2}{*}{500} & Data      & 3.72 & 3.76 & 2.77 & 2.78 & 2.93  \\
%                     & Data\&PDE & \pmb{3.48} & \pmb{3.50} & \pmb{2.37} & \pmb{2.45} & \pmb{2.66}  \\ 
% \hline
% \multirow{2}{*}{1000}& Data      & 3.11 & 3.19 & 1.99 & 2.01 & 2.29  \\
%                     & Data\&PDE & \pmb{2.96} & \pmb{2.98} & \pmb{1.82} & \pmb{1.78} & \pmb{2.09}  \\ 
% \hline
% \multirow{2}{*}{2000}& Data      & 2.62 & 2.65 & 1.55 & 1.61 & 1.72  \\
%                     & Data\&PDE & \pmb{2.49} & \pmb{2.51} & \pmb{1.41} & \pmb{1.40} & \pmb{1.53}  \\ 
% \hline
% multirow{2}{*}{5000}& Data      & 2.28 & 2.41 & 1.19 & 1.25 & 1.41  \\
%                     & Data\&PDE & \pmb{2.17} & \pmb{2.30} & \pmb{1.11} & \pmb{1.12} & \pmb{1.22}  \\
% \bottomrule
% \end{tabular}
% \end{table}

\begin{table}[htbp]
\caption{MRE of training strategies on dataset with different scales}\label{table3}
\centering
\setlength{\tabcolsep}{1.6mm}{
\begin{tabular}{cccccccccccc} 
\toprule
\multirow{2}{*}{Scale} & \multirow{2}{*}{Method} & \multicolumn{5}{c}{$MRE/\%$} & \multicolumn{5}{c}{$\sigma_{MRE}$}  \\ 
\cline{3-12}
                     &           & $u_{x}$ & $u_{y}$ & $\sigma_{xx}$ & $\sigma_{yy}$ & $\sigma_{xy}$ & $u_{x}$ & $u_{y}$ & $\sigma_{xx}$ & $\sigma_{yy}$ & $\sigma_{xy}$           \\ 
\hline
\multirow{2}{*}{200} & Data      & 6.46 & 6.42 & 4.43 & 4.55 & 5.52          &$\pm$0.0421  & $\pm$0.0396   &$\pm$0.0145     &$\pm$0.0143    &$\pm$0.0200              \\
                     & PDE &  \textbf{6.07} & \textbf{6.02} & \textbf{4.02}  & \textbf{4.17} & \textbf{5.11}          & \textbf{$\pm$0.0305} & \textbf{$\pm$0.0297} & \textbf{$\pm$0.0103}  & \textbf{$\pm$0.0102} & \textbf{$\pm$0.0142}           \\ 
\hline
\multirow{2}{*}{500} & Data      & 3.72 & 3.76 & 2.77 & 2.78 & 2.93           &$\pm$0.0185   &$\pm$0.0193   &$\pm$0.0083   &$\pm$0.0088   &$\pm$0.0104             \\
                     & PDE &  \textbf{3.48} & \textbf{3.50} & \textbf{2.37} & \textbf{2.45} & \textbf{2.66}          & \textbf{$\pm$0.0095} & \textbf{$\pm$0.0096} & \textbf{$\pm$0.0065} & \textbf{$\pm$0.0056} & \textbf{$\pm$0.0093}           \\ 
\hline
\multirow{2}{*}{1000}& Data      & 3.11 & 3.19 & 1.99 & 2.01 & 2.29           &$\pm$0.0114   &$\pm$0.0115   &$\pm$0.0056   &$\pm$0.0057   &$\pm$0.0086\\
                     & PDE & \textbf{2.96} & \textbf{2.98} & \textbf{1.82} & \textbf{1.78} & \textbf{2.09}          & \textbf{$\pm$0.0073} & \textbf{$\pm$0.0073} & \textbf{$\pm$0.0046} & \textbf{$\pm$0.0044} & \textbf{$\pm$0.0071}           \\ 
\hline
\multirow{2}{*}{2000}& Data       & 2.62 & 2.65 & 1.55 & 1.61 & 1.72            &$\pm$0.0089   &$\pm$0.0092   &$\pm$0.0049   &$\pm$0.0052   &$\pm$0.0065            \\
                     & PDE & \textbf{2.49} & \textbf{2.51} & \textbf{1.41} & \textbf{1.40} & \textbf{1.53}          & \textbf{$\pm$0.0063} & \textbf{$\pm$0.0062} & \textbf{$\pm$0.0044} & \textbf{$\pm$0.0042} & \textbf{$\pm$0.0044}           \\ 
\hline
\multirow{2}{*}{5000}& Data      & 2.28 & 2.41 & 1.19 & 1.25 & 1.41           &$\pm$0.0057   &$\pm$0.0069   &$\pm$0.0039   &$\pm$0.0040   &$\pm$0.0046        \\
                     & PDE & \textbf{2.17} & \textbf{2.30} & \textbf{1.11} & \textbf{1.12} & \textbf{1.22}          & \textbf{$\pm$0.0048} & \textbf{$\pm$0.0052} & \textbf{$\pm$0.0029} & \textbf{$\pm$0.0030} & \textbf{$\pm$0.0031}           \\
\bottomrule
\end{tabular}}
\end{table}

According to the experimental results, the physics-informed surrogate model exhibits better performance than data-driven only in data set with different sample sizes. Relatively speaking, the physics-informed strategy improves more significantly in the case of small samples, and the gap between the data-driven strategy and the physics-informed strategy gradually narrows as the number of samples increases, which demonstrates the potential of the physics-informed strategy in cases with small sample size.

\begin{figure*}[htbp]
	\centering
	{\includegraphics[width=0.95\linewidth]{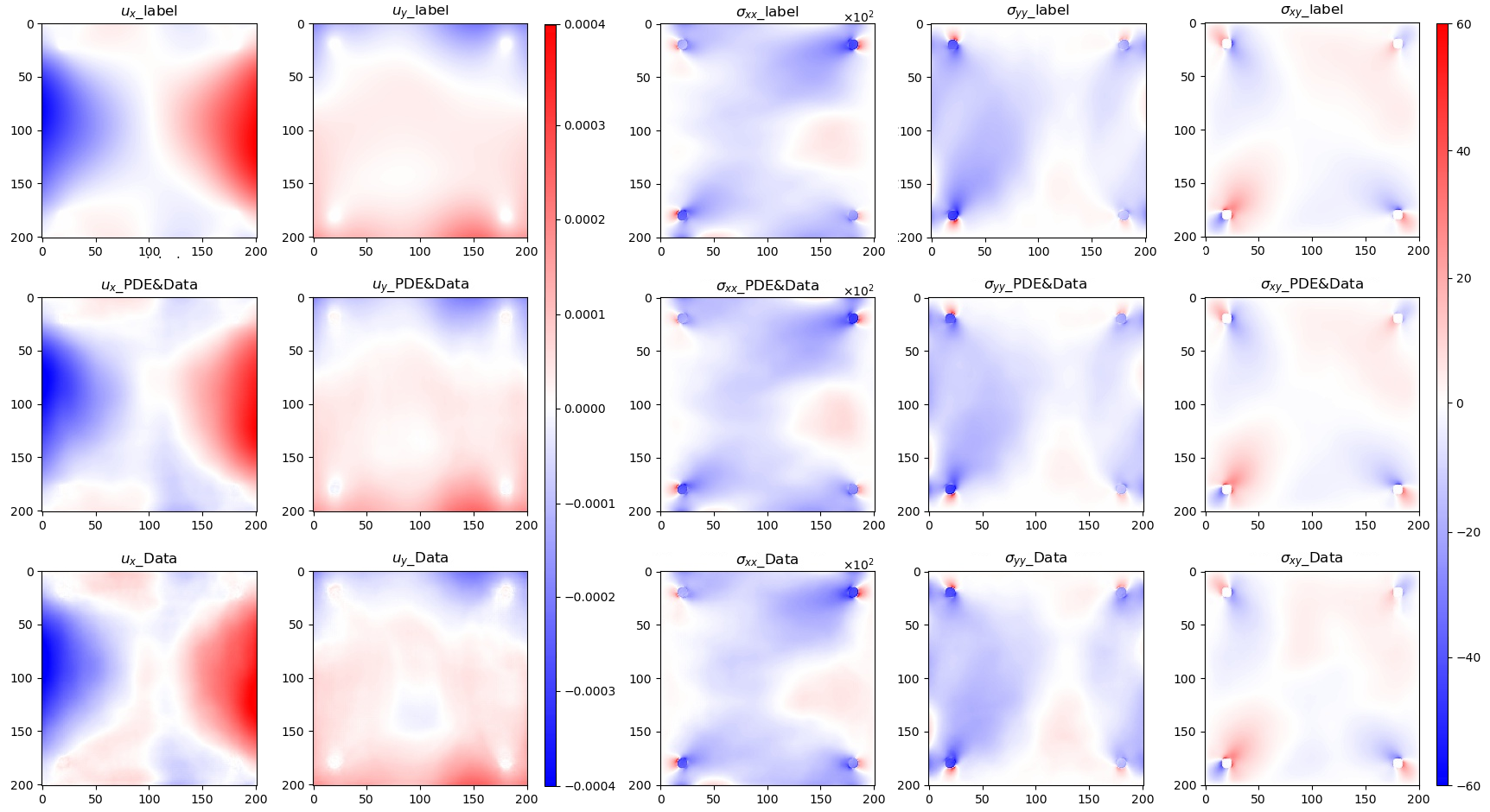}}
	\caption{Predictions of the data-driven strategy and the physics-informed strategy.}\label{result2}
\end{figure*}

In addition, we conduct training on smaller size of samples. Fig.\ref{result2} shows the prediction obtained by the data-driven strategy and the physics-informed strategy respectively for 100 samples. The comparison shows that physics-informed strategy obtains predictions that are more in line with the practical meaning of physics and realistic working conditions.

\section{Conclusion}\label{sec5}
To solve the regression problems of multiple physical quantities in thermal-mechanical coupled fields, this paper proposes a novel MTL network MTA-UNet. The MTA-UNet integrates the strengths of U-Net and MTL, having the advantage of selective sharing parameters of different tasks and layers. With shared encoding layers, task-specific decoding layers, and AG feature filtering operations of MTA-UNet, the correlation between different tasks is fully learned. The performance of the deep learning-based surrogate model is greatly enhanced, and the training time and system memory are also effectively decreased. Subsequently, an uncertainty-based dynamic loss weighted strategy is used among loss functions of different tasks. With this strategy, the training speed and accuracy of each sub task are well balanced. In addition, a physics-informed method is applied during training to rapidly predict the thermal stress and thermal deformation of satellite motherboards. In the process, a set of PDEs are encoded into a loss function, making full use of the prior physics knowledge. Compared with the data-driven only strategy, the physics-informed method exhibits better performance in training sets with different sample sizes and obtains solutions more consistent with the laws of physics.

Despite the MTA-UNet neural network we constructed for the regression problems of multiple physical quantities, further research is still required. As our model structure is strongly general, it is worth exploring strategies to enhance the balance between multiple tasks further. Extension to more complex boundaries and more complicated nonlinear problems is also valuable. Besides, embedding physics knowledge into the network structure remains a challenging work.

\section*{Conflict of interest statement}
On behalf of all authors, the corresponding author states that there is no conflict of interest.

\section*{Acknowledgments}
This work was supported by the National Natural Science Foundation of China (Nos.11725211 and 52005505).

\bibliography{mybib}

\end{document}